\documentclass[letterpaper,journal]{IEEEtran}
\usepackage{amsmath,amsfonts}
\usepackage{algorithmic}
\usepackage{algorithm}
\usepackage{array}
\usepackage[caption=false,font=normalsize,labelfont=sf,textfont=sf]{subfig}
\usepackage{textcomp}
\usepackage{stfloats}
\usepackage{url}
\usepackage{verbatim}
\usepackage{graphicx}
\usepackage{cite}
\begin{document}
\bstctlcite{References:BSTcontrol}

\title{DeltaWorld: Physically Consistent Interactive World Simulators via Action-Conditioned Latent Increment Learning}

\author{Boyuan Hou$^{1,\dagger}$, Xiaoge Cao$^{2,4,\dagger}$, Chaofan Zhang$^{3,*}$, Shuo Wang$^{3}$, and Shaowei Cui$^{1,3,*}$\\[0.5ex]
{\normalsize $^{1}$Imprintx Robotics}\\
{\normalsize $^{2}$Electric Power Research Institute of State Grid Jibei Electric Power Co., Ltd.}\\
{\normalsize $^{3}$Institute of Automation, Chinese Academy of Sciences}\\%
{\normalsize $^{4}$North China Electric Power Research Institute Co., Ltd.}
\thanks{$^\dagger$The two authors contributed equally to this work.}
\thanks{*Corresponding author: Chaofan Zhang and Shaowei Cui.}
}



\maketitle

\begin{abstract}
    Interactive world simulators can provide scalable environments for robot planning, policy training, and evaluation by predicting action consequences while reducing reliance on repeated physical rollouts. To serve these applications, they must generate future image sequences that respond faithfully to robot actions and preserve the dynamics of robot-object interactions over long horizons. However, existing world models typically predict the entire next latent state and often fail to capture subtle changes induced by robot actions. Such omissions can produce physically implausible outcomes, including object interpenetration and excessive deformation. To address this limitation, we propose DeltaWorld, a physically consistent interactive world simulator for robotic manipulation. Our method introduces the Delta Latent Transition Model (Delta-LTM), which predicts action-induced latent feature changes and adds them to the current latent state to obtain the next state, rather than predicting the next latent state directly. To mitigate object interpenetration and excessive deformation in predicted future frames, Interaction-aware Latent Alignment is introduced to construct counterfactual interaction regions and supervise interaction-related latent changes. DeltaWorld is evaluated on the IWS manipulation benchmark and a self-collected cross-robot dataset covering multiple robot embodiments and manipulation tasks. On the cross-robot dataset, DeltaWorld reduces FVD by 46.6\% and LPIPS by 31.1\% relative to the IWS baseline. These results highlight the potential of DeltaWorld for long-horizon action-conditioned video prediction in robotic manipulation. 
\end{abstract}

\begin{IEEEkeywords}
    World models, robot manipulation, latent dynamics, action-conditioned video prediction, physical consistency.
\end{IEEEkeywords}

\section{Introduction}
\IEEEPARstart{W}{ith} the increasing adoption of data-driven approaches in robotic manipulation, real-world robot interaction data have become a fundamental resource for learning complex manipulation behaviors~\cite{embodimentcollaboration2025openxembodimentroboticlearning,lin2026datascalinglawsimitation,zhaxizhuoma2025fastumiscalablehardwareindependentuniversal,zhao2026axisgrowablecommunitydrivendata,luo2026beingh05scalinghumancentricrobot}. However, policy training, action exploration, and performance evaluation often require repeated physical robot rollouts. These rollouts consume time and hardware resources, limiting the scale and reproducibility of experiments~\cite{embodimentcollaboration2025openxembodimentroboticlearning,wang2026interactiveworldsimulatorrobot}. World models offer a promising way to reduce reliance on physical rollouts by learning action-conditioned environment dynamics from existing robot interaction data. Such models can simulate the outcomes of robot actions, thereby supporting planning~\cite{zhou2025dinowmworldmodelspretrained,chen2026largevideoplannerenables}, policy learning, and evaluation~\cite{hafner2024masteringdiversedomainsworld,wang2026interactiveworldsimulatorrobot,guo2026ctrlworldcontrollablegenerativeworld}. Unlike conventional image prediction models that primarily focus on visual appearance changes, world models for robotic manipulation place greater emphasis on action-conditioned environment evolution, including object motion, robot--object interactions, and long-horizon action consequences~\cite{zhou2024robodreamerlearningcompositionalworld}.

To realize action-conditioned prediction, recent work has explored different representations of future robot observations. JEPA-style models forecast learned visual features without directly reconstructing pixels~\cite{assran2025vjepa2selfsupervisedvideo,ma2026feelworldvisuotactileworldmodel}. The Interactive World Simulator (IWS) learns action-conditioned dynamics in a compact image latent space and decodes predicted states into video~\cite{wang2026interactiveworldsimulatorrobot}. In parallel, video-generative world models support interactive evaluation by conditioning future frames on policy-derived latent actions~\cite{li2025worldevalworldmodelrealworld} or on frame-level controls with pose-conditioned memory~\cite{guo2026ctrlworldcontrollablegenerativeworld}. More recently, dWorldEval has modeled observations and actions jointly as discrete tokens, with sparse keyframes to support long-horizon prediction~\cite{li2026dworldevalscalableroboticpolicy}.

Despite these advances, predicting robot manipulation videos remains difficult when an action changes only a small part of the scene. Most of the image may stay the same between adjacent frames, while a robot action can move an object or alter its shape. Models that directly predict future frames or their latent representations may capture the largely unchanged scene while missing small changes to manipulated objects. The resulting errors can grow during autoregressive rollout and cause object drift or implausible deformation. Therefore, the key challenge is to predict the changes caused by robot actions, especially in regions where the robot interacts with objects.

To address this challenge, we propose DeltaWorld, a physically consistent interactive world simulator that predicts action-induced latent changes instead of the entire next latent state. Our method introduces the Delta Latent Transition Model (Delta-LTM), which estimates these changes and adds them to the current latent state to obtain the next state. Interaction-aware Latent Alignment uses counterfactual interaction regions to emphasize changes around robot--object interactions during training. Experiments on the IWS benchmark and a self-collected cross-robot dataset show improved action-conditioned video prediction quality and long-horizon consistency across multiple manipulation tasks and robot embodiments. Our main contributions are summarized as follows:

\begin{enumerate}
    \item We propose DeltaWorld, an interactive world simulator built on the Delta Latent Transition Model (Delta-LTM), which predicts action-conditioned latent increments to update the current state, promoting stable long-horizon prediction and physically plausible interaction dynamics.

    \item We introduce Interaction-aware Latent Alignment, which provides structured supervision over task-relevant regions and encourages the simulator to capture the physical regularities underlying object changes rather than appearance-dominated latent transitions.

    \item We conduct systematic evaluations on the IWS manipulation benchmark and a self-collected dataset covering multiple robot embodiments. Experimental results show that DeltaWorld outperforms existing world model baselines across image- and video-level metrics for action-conditioned video prediction, while ablation studies further verify the complementary roles of Delta-LTM and Interaction-aware Latent Alignment.
\end{enumerate}

\section{RELATED WORKS}

\subsection{World Models for Robotic Manipulation}
World models have recently emerged as an important research direction in robot learning. By learning predictive models of how the environment evolves under robot actions, robots can internally simulate possible future outcomes, enabling applications such as planning, control, policy steering, and policy evaluation~\cite{zhou2025dinowmworldmodelspretrained,chen2026largevideoplannerenables,hafner2024masteringdiversedomainsworld,wang2026interactiveworldsimulatorrobot,guo2026ctrlworldcontrollablegenerativeworld}. Early and classical world-modeling approaches typically learn environment dynamics in compact state spaces and use the learned models for model-based decision making and control~\cite{hafner2019learninglatentdynamicsplanning,hafner2020dreamcontrollearningbehaviors}. More recently, advances in visual generative modeling have motivated world models that learn action-conditioned environment dynamics directly from high-dimensional visual observations, often through compact latent representations~\cite{baldassarre2025featuresdinofoundationvideo,zhou2025dinowmworldmodelspretrained,wang2026interactiveworldsimulatorrobot}.

Existing visual world models have mainly developed along two representative directions. One line of work leverages visual priors learned by large-scale video generation models to support robotic planning and interaction through future video or visual trajectory generation~\cite{nvidia2026cosmos3omnimodalworld,jang2025dreamgenunlockinggeneralizationrobot,wan2025wanopenadvancedlargescale,chen2026largevideoplannerenables}. Another line builds action-conditioned predictive models directly from robot interaction data, allowing generated observations to explicitly respond to robot actions and further support interactive rollouts, policy steering, and policy evaluation~\cite{wang2026interactiveworldsimulatorrobot,guo2026ctrlworldcontrollablegenerativeworld,shang2026worldarenaunifiedbenchmarkevaluating}. These studies indicate that robotic world models are gradually evolving from open-ended visual generation toward interactive environment models that can capture action consequences and interaction outcomes. Nevertheless, effectively modeling action-driven environment state changes in a compact representation space remains a key challenge for achieving stable long-horizon prediction and physically consistent interaction.

\subsection{Action-conditioned Latent Dynamics}
To avoid long-horizon prediction directly in high-dimensional pixel space, an increasing number of recent studies have explored action-conditioned dynamics in compact latent spaces. These methods typically map visual observations into latent states using pretrained visual encoders, autoencoders, or joint-embedding models, and predict future states from historical latent representations and robot actions~\cite{zhou2025dinowmworldmodelspretrained,wang2026interactiveworldsimulatorrobot,maes2026leworldmodelstableendtoendjointembedding,garrido2026learninglatentactionworld}. For example, DINO-WM~\cite{zhou2025dinowmworldmodelspretrained} directly predicts future patch-level representations based on pretrained visual features, thereby avoiding explicit pixel reconstruction. IWS~\cite{wang2026interactiveworldsimulatorrobot} adopts a two-stage framework, where an action-conditioned latent dynamics model is learned after freezing the visual autoencoder.

Recent studies have further investigated whether latent dynamics representations truly preserve information relevant to control and interaction. One line of work employs joint-embedding prediction~\cite{lanji2026unijepaunifiedjointembeddingpredictive} or task-relevant representation learning~\cite{assran2025vjepa2selfsupervisedvideo} to encourage future latent representations to focus more on state information useful for planning~\cite{toso2026learninginvariantvisualrepresentations}. Another line of work further leverages predicted latent states, latent subgoals, or action-relevant representations for action generation and policy learning~\cite{chen2026lawamlatentworldaction,lei2026delew05inferringactionfuture,han2026enhancingpolicylearningworldaction,liu2026stagewamjointembeddingstageprediction,yang2026lilawamlightweightlatentreasoning}. These studies suggest that, compared with directly generating future observations at the pixel level, modeling action-conditioned environment evolution in latent space~\cite{huang2026learningtransferabledynamicspriors} can provide more compact and decision-relevant dynamics representations. However, most existing methods still take future latent states or future visual representations as their primary prediction targets, while explicitly modeling latent state transitions induced by robot actions has received comparatively less attention. In contrast, we focus on the action-conditioned latent transition itself and explicitly model this transition in interaction-relevant regions.

\subsection{World Models for Policy Training and Evaluation}
Beyond predicting future environment evolution, world models can also serve as interactive surrogate environments that provide scalable data sources for robot policy training. Imitation learning and vision-language-action policies typically rely on large amounts of high-quality real-robot demonstrations, while collecting such data is costly, hardware-dependent, and difficult to scale rapidly. To alleviate this bottleneck, recent studies have begun to use world models as interactive environments for data generation and policy optimization, supporting synthetic interaction trajectory generation, data augmentation, and world-model-based policy training and post-training~\cite{xiao2026worldenvleveragingworldmodel,zhu2025wmpoworldmodelbasedpolicy,liu2026worldvlaloopclosedlooplearningvideo,yang2026riseselfimprovingrobotpolicy,zhang2026practicalworldmodelbasedreinforcement,yu2026wmdaggerenablingefficientdata,sharma2026worldgymnasttrainingrobotsreinforcement}. IWS further demonstrates that demonstrations can be collected directly within a learned interactive world model to support the training of multiple imitation learning policies without requiring additional real-robot interaction~\cite{wang2026interactiveworldsimulatorrobot}.

Another important application of world models is robot policy evaluation. Real-world evaluation often requires repeated environment resets, human supervision, and substantial hardware runtime, while strictly controlling initial conditions across experiments is difficult. These factors make large-scale checkpoint selection and policy comparison challenging. To address this issue, an increasing number of studies have explored action-conditioned video world models as surrogates for the real environment, estimating policy performance through closed-loop interaction between the policy and the generated environment~\cite{li2025worldevalworldmodelrealworld,quevedo2025worldgymworldmodelenvironment,tseng2025scalablepolicyevaluationvideo,geminiroboticsteam2026evaluatinggeminiroboticspolicies,li2026dworldevalscalableroboticpolicy}. Works such as WorldEval~\cite{li2025worldevalworldmodelrealworld}, WorldGym~\cite{quevedo2025worldgymworldmodelenvironment}, and dWorldEval~\cite{li2026dworldevalscalableroboticpolicy} suggest that world-model-based evaluation can, to some extent, preserve the relative performance ordering among different policies or model checkpoints, while providing a scalable and reproducible environment for screening policies before real-world deployment.

\section{Method}

\subsection{Overview}
We propose a physically consistent interactive world simulator that learns action-conditioned environment evolution through structured latent-state transition modeling. Given a robot interaction sequence
\[
\mathcal{E}=\{(o_t,a_t)\}_{t=0}^{T},
\]
where \(o_t \in \mathbb{R}^{3\times H\times W}\) denotes the visual observation at time \(t\), and \(a_t \in \mathbb{R}^{d_a}\) denotes the corresponding robot action. Our goal is to predict future visual observations conditioned on an initial visual context and a sequence of robot actions.

\begin{figure*}[!t]
\centering
\includegraphics[width=\textwidth]{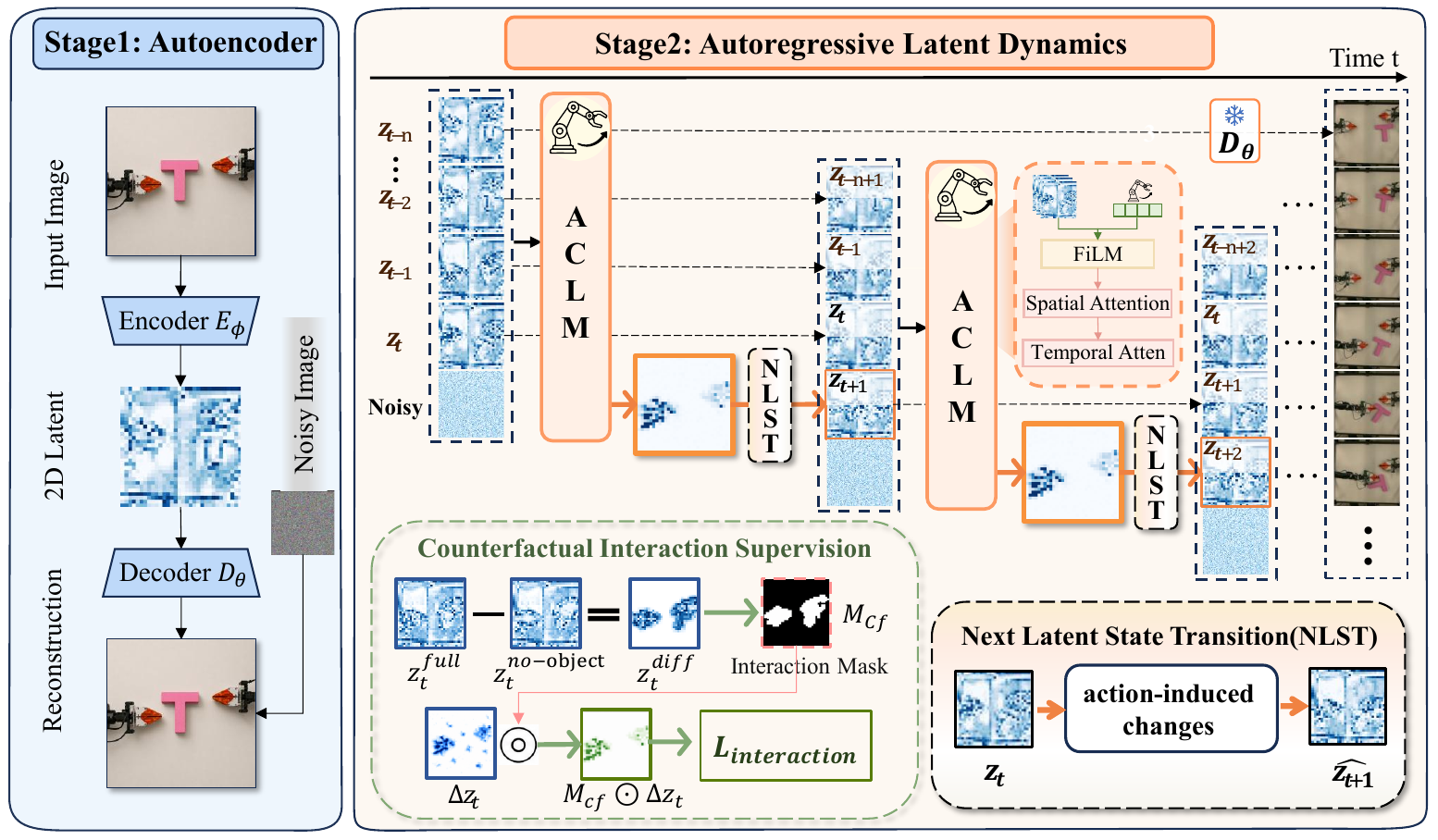}
\caption{Overview of the proposed physically consistent interactive world simulator. Stage~1 learns a compact 2D latent representation with an autoencoder. Stage~2 predicts action-induced latent changes with an autoregressive dynamics model. Comparing latent features of the original scene with those of a version lacking the interacting robot arm and object produces a mask that increases training weight near robot-object interactions.}
\label{fig:method_overview}
\end{figure*}

As shown in Fig.~\ref{fig:method_overview}, our framework follows a two-stage training paradigm. In the first stage, an encoder \(E_{\phi}\) maps each visual observation to a compact 2D latent representation:
\[
z_t = E_{\phi}(o_t), \qquad
z_t \in \mathbb{R}^{C\times H'\times W'}.
\]
We employ a consistency-based conditional decoder \(D_{\theta}\) to reconstruct each visual observation. To improve its robustness to perturbations in the latent representation, we inject small Gaussian noise into the encoded latent state during training:
\[
\tilde{z}_t = z_t + \epsilon,\qquad
\epsilon\sim\mathcal{N}(0,\sigma_z^2).
\]
Conditioned on the perturbed latent, the decoder maps a noisy observation to one at a lower noise level:
\[
\hat{o}_{t}^{\sigma'}
=
D_{\theta}\!\left(o_{t}^{\sigma};\sigma,\sigma',\tilde{z}_t\right),
\qquad \sigma>\sigma'\geq 0,
\]
where \(o_t^{\sigma}\) denotes the observation \(o_t\) corrupted at noise level \(\sigma\), and \(\sigma'\) denotes the target noise level. The clean observation corresponds to \(\sigma'=0\). This stage aims to learn compact latent representations that retain sufficient visual information while improving the decoder's tolerance to small latent prediction errors from the dynamics model, thereby providing a stable state space for subsequent action-conditioned dynamics modeling.

In Stage~2, the encoder and decoder are frozen while an action-conditioned dynamics model is trained to predict the latent feature increment induced by robot actions. Given the latest $N$ latent states and the corresponding action sequence, the model estimates the change from the current to the next latent state rather than generating the entire next state. The predicted increment is added to the current state, and this update is repeated autoregressively to produce future latent states. By carrying largely unchanged scene features forward, the residual update focuses learning on action-driven object motion and robot-object interactions.

Although Delta-LTM makes action-conditioned latent increments the primary prediction target, not all latent changes are equally relevant to robot-object interaction. Changes related to robot interaction typically occupy only a small portion of the latent representation, while most latent locations mainly encode static objects or background appearance. Therefore, applying uniform supervision across all latent locations may weaken the learning signals associated with robot--object interactions. To address this issue, we further propose Interaction-aware Latent Alignment. Specifically, we construct a counterfactual interaction map by comparing the latent representations of the original observation and a corresponding counterfactual observation in which the manipulated object and the robot arm regions involved in the current interaction are removed. This interaction map identifies latent regions that are sensitive to robot--object interactions and assigns greater weight to the corresponding state transitions during dynamics learning.

During inference, the simulator starts from observed latent states and autoregressively predicts future latent states conditioned on a sequence of robot actions. Each predicted latent state is decoded by the frozen decoder to generate the corresponding RGB observation, which is then used as context for subsequent predictions. In this way, DeltaWorld supports long-horizon interactive rollouts under robot action conditioning and provides an interactive predictive environment for downstream robotic world-model applications.

\subsection{Delta Latent Transition Model}
Action-conditioned world models aim to capture how robot actions transform the current environment state into the next one. We therefore formulate the action-induced latent increment as the explicit prediction target of the dynamics model, rather than treating it implicitly within complete future-state prediction. This formulation is particularly suitable for robotic manipulation: much of the visual scene changes little between adjacent frames and the action-induced dynamics are reflected in robot motion, object displacement, contact changes, and interaction responses. We introduce the Delta Latent Transition Model (Delta-LTM), which estimates the latent increment from the current state and robot actions and applies it to update the current state. In this way, the model learns how robot actions change the latent state rather than directly predicting the complete future state.

Given two consecutive latent states $z_t$ and $z_{t+1}$, we define their latent-space feature increment as
\begin{equation}
\Delta z_t = z_{t+1}-z_t .
\end{equation}
Rather than directly predicting the entire latent state at the next time step, Delta-LTM predicts the action-conditioned latent increment from the latest $N$ latent states and the corresponding action sequence:
\begin{equation}
    \widehat{\Delta z}_t =
    F_{\psi}\left(
    z_{t-N+1:t},
    a_{t-N+1:t}
    \right),
    \end{equation}
where $F_{\psi}$ denotes the action-conditioned latent dynamics model, and $N$ is the history context length. The predicted latent state at the next time step is then obtained through a residual state update:
\begin{equation}
\hat z_{t+1}
=
z_t+\widehat{\Delta z}_t .
\end{equation}
By defining \(\Delta z_t\) as the explicit prediction target, Delta-LTM directly associates action conditioning with the latent state change it produces. The residual update carries persistent information from \(z_t\) into the next state, allowing the dynamics model to concentrate its capacity on predicting action-conditioned evolution.

To preserve the ability to model complex spatiotemporal dynamics in robotic interaction, we adopt an action-conditioned consistency dynamics model as the underlying architecture of \(F_{\psi}\). The model represents the latent sequence as a spatiotemporal tensor and uses 3D convolutions to extract local spatiotemporal features. Robot actions are injected through FiLM conditioning, while spatial attention and temporal attention are employed to model dependencies across spatial regions and temporal dimensions. Accordingly, \(F_{\psi}\) predicts the action-conditioned latent transition as its output, while the next latent state is recovered through the residual update in Eq.~(3).

To supervise the predicted latent transition, we constrain the residual between \(\widehat{\Delta z}_t\) and the ground-truth transition \(\Delta z_t\). This residual serves as the basis for both global dynamics supervision and the subsequent interaction-aware local supervision. The specific training objectives are presented together in Sec.~\ref{sec:training_inference}.


\subsection{Interaction-aware Latent Alignment}

In robotic manipulation, critical state changes typically occur around robot-object interactions, while the background remains relatively stable. When every latent location contributes equally to the training loss, the many background locations can overshadow the few locations where the robot interacts with objects. The model may then miss small changes in object position or contact. To address this issue, we introduce Interaction-aware Latent Alignment, which uses counterfactual interaction supervision to identify latent regions associated with robot-object interactions and assigns greater weight to prediction errors in these regions.

Specifically, given the original observation $o_t^{\mathrm{full}}$ at time $t$, we first obtain the latent state of the complete scene using the frozen encoder:
\begin{equation}
z_t^{\mathrm{full}}
=
E_{\phi}\left(o_t^{\mathrm{full}}\right).
\end{equation}
We then construct a corresponding counterfactual observation $o_t^{\mathrm{cf}}$ by jointly removing the manipulated object and the robot arm regions involved in the current interaction, while preserving the remaining scene content. Using the same encoder, we obtain
\begin{equation}
z_t^{\mathrm{cf}}
=
E_{\phi}\left(o_t^{\mathrm{cf}}\right).
\end{equation}

The two observations differ mainly in the manipulated object and the robot arm involved in the interaction. A large difference between their encoded features at a latent location therefore suggests that this location is associated with the interaction. We define the counterfactual latent difference as
\begin{equation}
z_t^{\mathrm{diff}}
=
\left|
z_t^{\mathrm{full}}
-
z_t^{\mathrm{cf}}
\right|.
\end{equation}

\begin{figure}[!t]
    \centering
    \includegraphics[width=\columnwidth]{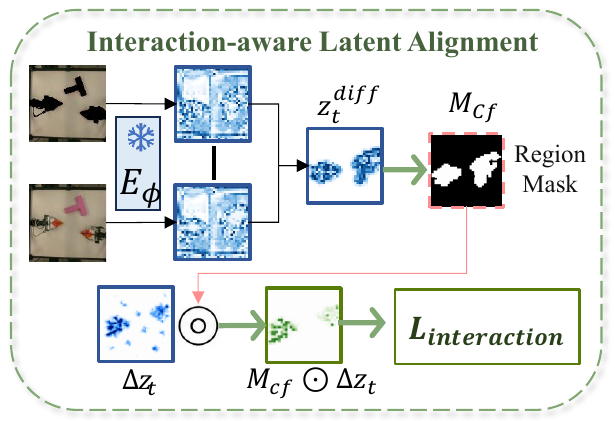}
    \caption{
    Illustration of Interaction-aware Latent Alignment. Comparing the latent representations of the original and interaction-removed observations yields a mask that emphasizes robot--object interaction regions during training.
    }
    \label{fig:interaction_alignment}
\end{figure}
We average the latent differences across channels, normalize the resulting map, and threshold it to obtain a binary interaction region mask:
\begin{equation}
    M_t^{\mathrm{int}}
    =
    \mathcal{B}
    \left(
    \mathcal{N}
    \left(
    \frac{1}{C}
    \sum_{c=1}^{C}
    z_{t,c}^{\mathrm{diff}}
    \right)
    \right),
\end{equation}
where $\mathcal{N}(\cdot)$ denotes the normalization operation and $\mathcal{B}(\cdot)$ denotes the thresholding operation used to obtain a binary interaction region mask. The resulting $M_t^{\mathrm{int}}$ is used to localize latent regions that are highly relevant to robot--object interactions. As shown in Fig.~\ref{fig:interaction_alignment}, Interaction-aware Latent Alignment first removes the robot--object interaction regions from the original observation and extracts latent representations from both the full and counterfactual observations using the same frozen encoder. The resulting latent difference is then used to construct the interaction region mask, which provides a spatial constraint for the subsequent localized supervision of latent state transitions.

Based on the interaction region mask, we further apply localized supervision to the latent increments modeled by Delta-LTM. Given the ground-truth latent increment $\Delta z_t$ and the predicted latent increment $\widehat{\Delta z}_t$, we first define the latent increment prediction residual as
\begin{equation}
    e_t
    =
    \widehat{\Delta z}_t
    -
    \Delta z_t .
\end{equation}
We then extract the interaction-region residual using the interaction mask:
\begin{equation}
    e_t^{\mathrm{int}}
    =
    M_t^{\mathrm{int}}
    \odot
    e_t ,
\end{equation}
where $\odot$ denotes element-wise multiplication, with the spatial mask broadcast along the latent channel dimension.

Interaction-aware Latent Alignment does not introduce an additional dynamics prediction target. Instead, it applies spatially selective supervision to the same latent increment prediction. The interaction region mask encourages the model to focus on latent changes associated with robot-object interactions. 

\subsection{Training and Inference}
\label{sec:training_inference}

After Stage~1, the encoder $E_{\phi}$ and decoder $D_{\theta}$ are frozen, and only the dynamics model $F_{\psi}$ is optimized. Ground-truth observations provide teacher-forced latent histories, while noise at level $s$ improves robustness to imperfect rollout contexts. The target increment and its prediction residual are
\begin{equation}
\Delta z_t^{s}=z_{t+1}^{s}-z_t^{s},
\qquad
e_t^{s}=\widehat{\Delta z}_t^{s}-\Delta z_t^{s}.
\end{equation}
The residual is penalized globally and within the interaction mask $M_t^{\mathrm{int}}$:
\begin{equation}
\mathcal{L}_{\mathrm{global},t}
=
\left\|e_t^{s}\right\|_2^2
+
\lambda_{\Delta}
\left\|e_t^{s}\right\|_1 .
\end{equation}
\begin{equation}
\begin{aligned}
    \mathcal{L}_{\mathrm{int},t}
    ={}&
    \lambda_{\mathrm{obj-z}}
    \left\|
    M_t^{\mathrm{int}}
    \odot
    e_t^{s}
    \right\|_2^2
    \\
    &+
    \lambda_{\mathrm{obj-\Delta}}
    \left\|
    M_t^{\mathrm{int}}
    \odot
    e_t^{s}
    \right\|_1 .
\end{aligned}
\end{equation}
The combined Stage~2 objective is
\begin{equation}
    \mathcal{L}_{\mathrm{dyn}}
    =
    \mathbb{E}_{t}
    \left[
    \omega_t
    \left(
    \mathcal{L}_{\mathrm{global},t}
    +
    \gamma(k)
    \mathcal{L}_{\mathrm{int},t}
    \right)
    \right],
\end{equation}
Here, $\omega_t$ emphasizes more distant prediction steps, and $\gamma(k)$ gradually increases the contribution of the interaction loss during training. At inference, the initial observation is encoded, and $F_{\psi}$ predicts each latent increment from a fixed-length history of latent states and the corresponding action sequence. Each increment updates the current latent state; the updated state enters the sliding history and is decoded into an RGB frame by $D_{\theta}$.

\section{EXPERIMENTS}
In this section, we evaluate DeltaWorld from three perspectives: 1) How does DeltaWorld compare with state-of-the-art world models in terms of action-conditioned video prediction quality and long-horizon interaction consistency? 2) Can the proposed latent increment formulation remain effective across different robot embodiments and manipulation tasks? 3) How does each proposed component contribute to the overall performance? We first introduce the experimental setup in Section IV-A. Section IV-B presents quantitative and qualitative comparisons on both the IWS benchmark and our self-collected cross-robot dataset, while Section IV-C provides ablation studies of Delta-LTM and Interaction-aware Latent Alignment.

\subsection{World Model Instantiation}
We instantiate DeltaWorld on five robotic manipulation tasks from the Interactive World Simulator (IWS) benchmark, including T-pushing, bimanual box manipulation, bimanual sweeping, bimanual rope manipulation, and single-arm grasping. These tasks cover rigid-object manipulation, deformable-object interaction, as well as both single-arm and bimanual manipulation settings. For each task, the training set contains approximately 500--600 interaction episodes. Each dataset provides two camera views. Following the experimental protocol of IWS, we train and evaluate independent world models for different camera views.

To further evaluate the applicability of DeltaWorld across different robot platforms, we additionally collect four manipulation tasks using three different robot platforms, including whiteboard wiping, chip grasping, cable arrangement, and Peg-in-Hole. The corresponding training sets contain 127, 329, 886, and 893 episodes, respectively, covering different robot embodiments, action spaces, and forms of robot--object interaction. Unlike the IWS datasets, the interaction trajectories in our self-collected datasets have task-dependent variable lengths, preserving the natural variation in trajectory duration that arises during real-world data collection across different manipulation tasks.

All visual observations are represented as RGB images with a resolution of ($128\times128$). Stage~1 autoencoder training uses individual image frames as training samples, while Stage~2 latent dynamics training uses sequence windows of 10 frames. During evaluation, the model takes the first observed frame as the initial condition and autoregressively generates future prediction sequences of up to 150 frames.

\subsection{Video Baseline Comparisons}
\begin{figure*}[!t]
    \centering
    \includegraphics[width=\textwidth]{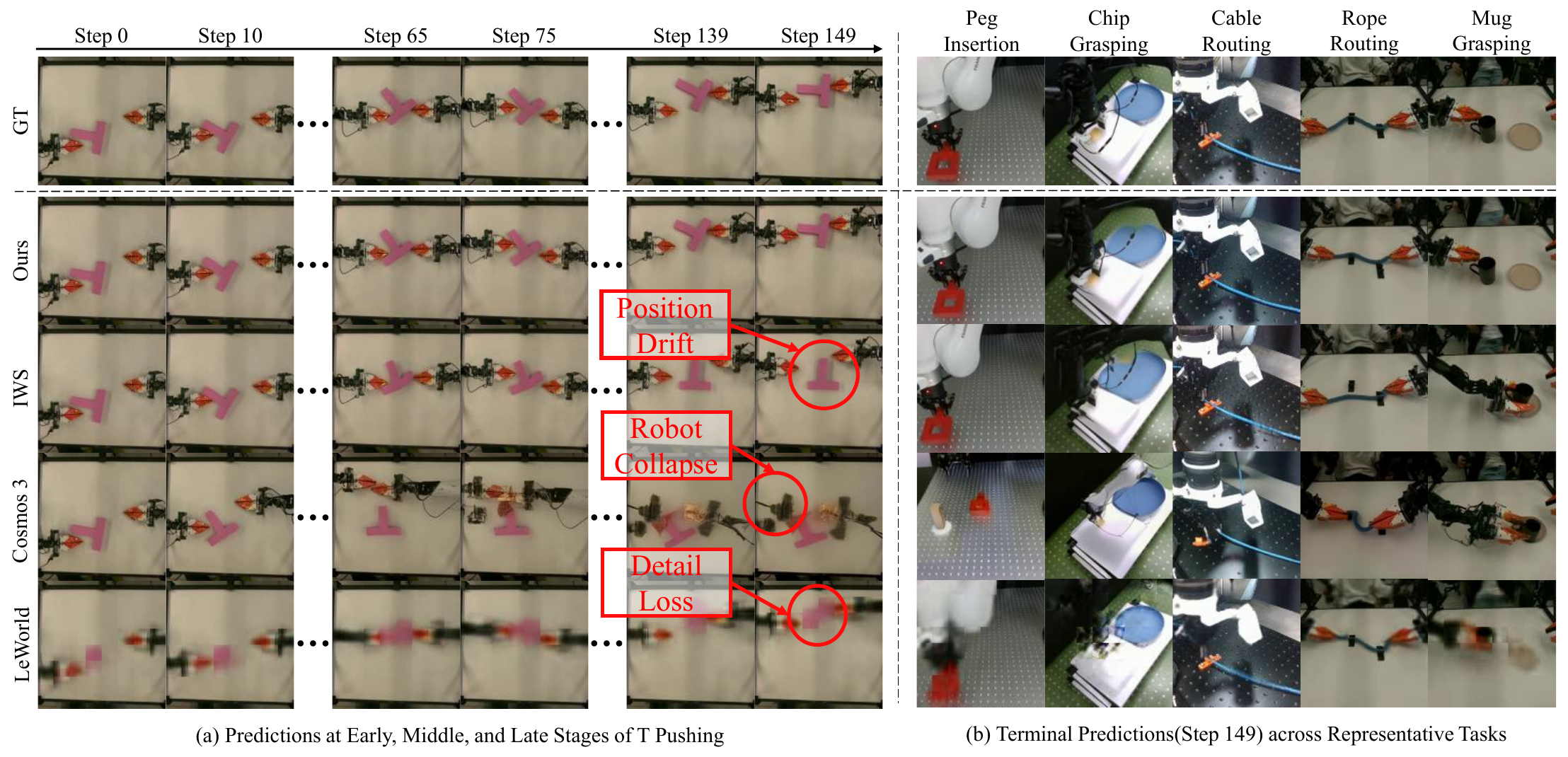}
    \caption{Qualitative comparison of long-horizon action-conditioned video prediction. (a) Predictions on the T Pushing task at the beginning, middle, and end of a 150-frame autoregressive rollout, with paired frames separated by 10 steps at each stage. Compared with the baselines, DeltaWorld better preserves object position, robot appearance, and fine-grained visual details over long-horizon prediction. (b) Terminal predictions at Step 149 across representative manipulation tasks, including Peg Insertion, Chip Grasping, and Cable Routing from our self-collected dataset, and Rope Routing and Mug Grasping from the IWS benchmark.}
    \label{fig:qualitative}
    \end{figure*}

To evaluate the performance of DeltaWorld on action-conditioned video prediction, we compare it with several representative world models, including LeWorld~\cite{maes2026leworldmodelstableendtoendjointembedding}, Cosmos3~\cite{nvidia2026cosmos3omnimodalworld}, and IWS~\cite{wang2026interactiveworldsimulatorrobot}. All methods use the same data splits and evaluation protocol. We report FID, FVD, LPIPS, PSNR, SSIM, and MSE to comprehensively evaluate visual quality, perceptual consistency, and temporal prediction performance.

For the baseline models, LeWorld and IWS are trained from scratch following their publicly released implementations and default training configurations, with the corresponding method-specific hyperparameter settings. Cosmos 3 provides an official fine-tuning pipeline for robotic data. We therefore first convert the robot interaction data into the required format and then fine-tune the model using the default Cosmos 3  fine-tuning settings. All methods are trained using the same training data and validation splits and are evaluated under a consistent rollout setting.

We first compare the methods on five robotic manipulation tasks from the IWS benchmark under two camera views. Table~\ref{tab:iws} summarizes the average results across all tasks and views. DeltaWorld achieves the best overall performance across all evaluation metrics. Compared with IWS, FID and FVD are reduced from 44.16 and 264.52 to 36.04 and 231.98, respectively, while LPIPS decreases from 0.0661 to 0.0490. PSNR, SSIM, and MSE also show consistent improvements. These results indicate that explicitly modeling action-driven latent state transitions improves perceptual and temporal consistency over long-horizon prediction while preserving visual reconstruction quality.

\begin{table}[!t]
\caption{Action-conditioned video prediction on the IWS manipulation benchmark, averaged over five tasks and two camera views.}
\label{tab:iws}
\centering
\footnotesize
\setlength{\tabcolsep}{2.4pt}
\renewcommand{\arraystretch}{1.15}
\begin{tabular}{@{}lcccccc@{}}
\hline
Method & FID $\downarrow$ & FVD $\downarrow$ & LPIPS $\downarrow$ & PSNR $\uparrow$ & SSIM $\uparrow$ & MSE $\downarrow$ \\
\hline
LeWorld~\cite{maes2026leworldmodelstableendtoendjointembedding} & 149.89 & 1647.68 & 0.1655 & 24.89 & 0.8551 & 0.0048 \\
Cosmos3~\cite{nvidia2026cosmos3omnimodalworld} & 75.38 & 612.39 & 0.2948 & 16.86 & 0.4091 & 0.0853 \\
IWS~\cite{wang2026interactiveworldsimulatorrobot} & 44.16 & 264.52 & 0.0661 & 28.80 & 0.8537 & 0.0054 \\
\textbf{Ours} & \textbf{36.04} & \textbf{231.98} & \textbf{0.0490} & \textbf{31.22} & \textbf{0.9011} & \textbf{0.0039} \\
\hline
\end{tabular}
\end{table}

To further examine the applicability of DeltaWorld across different robot embodiments, we conduct the same comparison on our self-collected cross-robot dataset, which includes three robot platforms: Imeta, UR5, and Franka. As shown in Table~\ref{tab:cross-robot}, DeltaWorld maintains consistent advantages across different robot structures, action spaces, and interaction patterns. Compared with IWS, FVD is substantially reduced from 168.98 to 90.31, corresponding to a relative improvement of approximately 46.6\%, while LPIPS, PSNR, SSIM, and MSE are also further improved. The improvement in FVD suggests that the main advantage of DeltaWorld lies in modeling action-conditioned dynamics and long-horizon temporal consistency, rather than merely improving single-frame appearance quality. These results further indicate that the proposed latent transition modeling approach can be applied consistently across different robot embodiments and manipulation tasks.

\begin{table}[!t]
\caption{Action-conditioned video prediction on our cross-robot manipulation dataset, averaged over four tasks across three robot platforms.}
\label{tab:cross-robot}
\centering
\footnotesize
\setlength{\tabcolsep}{2.4pt}
\renewcommand{\arraystretch}{1.15}
\begin{tabular}{@{}lcccccc@{}}
\hline
Method & FID $\downarrow$ & FVD $\downarrow$ & LPIPS $\downarrow$ & PSNR $\uparrow$ & SSIM $\uparrow$ & MSE $\downarrow$ \\
\hline
LeWorld~\cite{maes2026leworldmodelstableendtoendjointembedding} & 150.30 & 1052.66 & 0.1161 & 21.24 & 0.8598 & 0.0091 \\
Cosmos3~\cite{nvidia2026cosmos3omnimodalworld} & 105.95 & 876.19 & 0.2349 & 14.79 & 0.5376 & 0.1495 \\
IWS~\cite{wang2026interactiveworldsimulatorrobot} & 39.37 & 168.98 & 0.0440 & 28.20 & 0.8700 & 0.0075 \\
\textbf{Ours} & \textbf{37.26} & \textbf{90.31} & \textbf{0.0303} & \textbf{30.21} & \textbf{0.9022} & \textbf{0.0047} \\
\hline
\end{tabular}
\end{table}

Fig.~\ref{fig:qualitative} presents qualitative prediction results of different methods on representative tasks, including both the IWS benchmark and our self-collected cross-robot tasks. Compared with the baseline models, DeltaWorld more consistently preserves object structure, robot motion, and robot--object interaction relationships over longer prediction horizons, while reducing visual drift and dynamic inconsistencies accumulated during autoregressive rollout.

\subsection{Ablation Study}
To analyze the contributions of the Delta Latent Transition Model (Delta-LTM) and Interaction-aware Latent Alignment to the overall model performance, we conduct ablation studies by removing one component at a time while keeping all other training and evaluation settings unchanged. Table~\ref{tab:ablation} summarizes the average results on the IWS benchmark and our self-collected cross-robot dataset.

\begin{table}[!t]
\caption{Ablation results averaged over the IWS benchmark and the self-collected cross-robot dataset.}
\label{tab:ablation}
\centering
\footnotesize
\setlength{\tabcolsep}{2.4pt}
\renewcommand{\arraystretch}{1.15}
\begin{tabular}{@{}>{\raggedright\arraybackslash}p{1.85cm}cccccc@{}}
\hline
Method & FID $\downarrow$ & FVD $\downarrow$ & LPIPS $\downarrow$ & PSNR $\uparrow$ & SSIM $\uparrow$ & MSE $\downarrow$ \\
\hline
w/o Int. Align. & 37.93 & 207.60 & 0.0525 & 29.73 & 0.8822 & 0.0052 \\
w/o Delta-LTM & 37.92 & 228.03 & 0.0569 & 29.17 & 0.8628 & 0.0063 \\
\textbf{DeltaWorld} & \textbf{36.39} & \textbf{191.50} & \textbf{0.0437} & \textbf{30.93} & \textbf{0.9014} & \textbf{0.0042} \\
\hline
\end{tabular}
\end{table}

The complete DeltaWorld achieves the best overall performance across all evaluation metrics. Removing Delta-LTM leads to clear degradation in FVD, LPIPS, PSNR, SSIM, and MSE. In particular, FVD increases from 191.50 to 228.03, indicating that directly predicting future absolute latent states weakens the model’s ability to capture long-horizon dynamics. Removing Interaction-aware Latent Alignment also degrades performance across the evaluated metrics, with consistent drops in FVD, LPIPS, and reconstruction-quality measures. This suggests that strengthening supervision over latent regions associated with robot–object interactions further improves perceptual consistency and interaction stability.

Delta-LTM reformulates the dynamics prediction target along the temporal dimension by estimating action-conditioned latent increments and applying them to the current state, whereas Interaction-aware Latent Alignment strengthens robot–object interaction-related state changes along the spatial dimension. Their combination yields the best overall performance, supporting the importance of both modeling what changes and emphasizing which changes matter for learning stable and interaction-consistent latent dynamics.

\section{Conclusion}
In this work, we presented DeltaWorld, a physically consistent interactive world simulator for robotic world models. Instead of treating the complete future latent state as the primary prediction target, the Delta Latent Transition Model (Delta-LTM) estimates action-conditioned latent increments and applies them to update the current state. Interaction-aware Latent Alignment further emphasizes transition errors in robot--object interaction-sensitive regions through counterfactual supervision. Together, these two components provide an increment-based and interaction-aware formulation for learning robotic world dynamics. Experiments on the IWS manipulation benchmark and a self-collected multi-robot dataset demonstrate improved action-conditioned video prediction across image- and video-level metrics, while ablation studies confirm the complementary roles of the two components.


\bibliographystyle{IEEEtran}
\bibliography{ref}

\newpage

\vfill

\end{document}